\documentclass[sigconf,nonacm]{acmart}
\AtBeginDocument{%
  }

\setcopyright{acmlicensed}
\copyrightyear{20xx}
\acmYear{20xx}
\acmDOI{XXXXXXX.XXXXXXX}
\acmConference[Conference acronym 'XX]{Make sure to enter the correct
  conference title from your rights confirmation email}{August xx--xx,
  20xx}{Chicago, IL}
\acmISBN{}

\renewcommand\footnotetextcopyrightpermission[1]{}
\usepackage{multirow}
\usepackage{tikz}
\usetikzlibrary{arrows.meta,positioning}
\usepackage{graphicx}

\usepackage{amsmath,amssymb}
\usepackage{xcolor}
\usepackage[T1]{fontenc}
\usepackage{caption}
\usepackage{subcaption}
 
\usetikzlibrary{
  arrows.meta,
  positioning,
  calc,
  fit,
  backgrounds,
  decorations.pathreplacing,
  decorations.markings,
  patterns,
  shapes.geometric,
  matrix
}
 
\definecolor{virpurple}{HTML}{534AB7}
\definecolor{virpurplelight}{HTML}{EEEDFE}
\definecolor{virteal}{HTML}{0F6E56}
\definecolor{virteallight}{HTML}{E1F5EE}
\definecolor{vircoral}{HTML}{993C1D}
\definecolor{vircorallight}{HTML}{FAECE7}
\definecolor{virblue}{HTML}{185FA5}
\definecolor{virbluelight}{HTML}{E6F1FB}
\definecolor{virgray}{HTML}{5F5E5A}
\definecolor{virgraylight}{HTML}{F1EFE8}
\definecolor{viramber}{HTML}{BA7517}
\definecolor{viramberlight}{HTML}{FAEEDA}
 
\tikzset{
  box/.style={
    draw=#1, fill=#1!8, rounded corners=4pt,
    minimum height=10mm, align=center, font=\small,
    line width=0.5pt
  },
  box/.default=virgray,
  bigbox/.style={
    draw=#1, fill=#1!6, rounded corners=8pt,
    minimum height=12mm, align=center, font=\small,
    line width=0.5pt
  },
  bigbox/.default=virgray,
  arr/.style={
    -{Stealth[length=5pt, width=4pt]},
    line width=0.6pt, color=virgray
  },
  darr/.style={
    -{Stealth[length=4pt, width=3pt]},
    line width=0.5pt, dashed, color=#1!60
  },
  darr/.default=virgray,
  lbl/.style={font=\scriptsize\sffamily, text=virgray},
  tlbl/.style={font=\scriptsize\sffamily\bfseries},
  plus/.style={
    circle, draw=virgray, inner sep=0pt,
    minimum size=7mm, font=\small\bfseries
  },
  frozen/.style={
    rounded corners=6pt, fill=#1!15,
    font=\tiny\sffamily, text=#1, inner sep=2pt
  },
  frozen/.default=virblue,
  trainable/.style={
    rounded corners=6pt, fill=virpurple!15,
    font=\tiny\sffamily, text=virpurple, inner sep=2pt
  },
}
 
\AtBeginDocument{\let\balance\relax}
\begin{document}

\title[HiRo: A Compact Four-Directional Hierarchical Reservoir Token-Mixer]{HiRo: A Compact Four-Directional Hierarchical Reservoir Token-Mixer for Efficient Image Classification}


\author{Md Farhadul Islam}
\affiliation{%
  \institution{University of Kentucky}
  \city{Lexington}
  \country{USA}}
\email{Farhadul.Islam@uky.edu}

\author{Ishan Thakkar}
\affiliation{%
  \institution{University of Kentucky}
  \city{Lexington}
  \country{USA}}
\email{igthakkar@uky.edu}

\author{J. Todd Hastings}
\affiliation{%
  \institution{University of Kentucky}
  \city{Lexington}
  \country{USA}}
\email{todd.hastings@uky.edu}

\begin{abstract}
Recent image classification models must balance local feature modeling, cross-window interaction, and parameter efficiency. Many high-performing architectures rely on fully trainable token-mixers, which improve representation learning but increase parameter count, optimization complexity and computational cost. We propose a parameter-efficient image classification model called HiRo that integrates shifted-window partitioning with multi-directional hierarchical reservoir computing. Images are divided into non-overlapping patches (treated as tokens), linearly projected, normalized, and enriched with 2D sinusoidal positional encodings, then processed within local windows. Inside each window, tokens are scanned in four directions and passed through a two-stage slice-and-mix reservoir module. In the first stage, directional sequences are split into contiguous slices, each processed by its own fixed reservoir with a trainable closed-loop readout. The resulting slice outputs are summarized using the start, end, and mean representations, and then mixed by a second-stage fixed reservoir for each direction. The mixed slice representations are expanded back to the token level and fused with the first-stage outputs, after which the four directional outputs are realigned and averaged. Consecutive blocks alternate between regular and shifted windows to enable cross-window interaction, followed by layer normalization, a residual feed-forward network, and global pooling for classification. This design combines regular and shifted window partitioning with hierarchical multi-directional reservoirs to make an efficient local-to-cross-window token-mixing framework for image classification. Despite using under 1M trainable parameters and significantly lower memory and time than transformer-style baselines, HiRo also achieves 99.46\%, 85.57\%, and 59.10\% accuracy on MNIST, CIFAR-10, and CIFAR-100, respectively.

\end{abstract}



\keywords{Reservoir Computing, Image Classification, Four-directional Scanning, Window Partitioning}


\maketitle

\section{Introduction}

Image classification models must balance representational power, computational efficiency, and inductive bias \cite{liu2024lightweight, wang2023theoretical, cetto2019size}. Convolutional models remain strong baselines because locality and parameter sharing are built into their design, whereas transformer-based vision models instead operate on patch tokens and use explicit token mixing for visual recognition. In particular, Vision Transformer showed that images can be processed as sequences of patches, and Swin Transformer introduced shifted window processing as an efficient mechanism for local modeling with cross-window interaction \cite{dosovitskiy2021image, liu2021swin}. Despite these advances, many high-performing visual backbones rely on fully trainable token mixing modules, which tend to increase parameter count, optimization complexity, and computational cost.

Reservoir computing offers a complementary perspective on sequence modeling. Instead of learning the full recurrent dynamics, reservoir methods keep the recurrent system fixed and train only the readout, allowing nonlinear temporal processing with reduced training cost \cite{jaeger2001echo, lukovsevivcius2009reservoir, lukovsevivcius2012practical}. However, despite this efficiency, reservoir computing has been explored less extensively in modern patch-based vision architectures, particularly in settings that demand both structured local spatial modeling and controlled communication across neighboring regions. This is also relevant from a hardware perspective, since fixed recurrent dynamics can be realized by physical or neuromorphic reservoirs, where the substrate performs nonlinear transformation while only lightweight readouts require training.

Motivated by this gap, we propose a windowed parallel reservoir architecture for image classification that brings reservoir computing principles into a patch-based vision setting. The model first extracts non-overlapping patches, projects them into an embedding space, applies normalization, and adds fixed two-dimensional sinusoidal positional embeddings. This design preserves the patch token structure of the image while encoding spatial location without introducing additional trainable parameters.

The resulting token grid is then processed by Windowed Parallel Reservoir blocks. Inside each local window, tokens are scanned in four deterministic directions, namely left to right, right to left, top to bottom, and bottom to top. This multi-directional scanning sees complementary spatial orderings that a single traversal would miss. 
Each directional sequence is passed through a hierarchical slice-mixing mechanism. In the first stage, the sequence is divided into equal contiguous slices, and each slice is processed by a distinct fixed sub-reservoir coupled to a trainable closed-loop readout. This arrangement shortens the effective recurrent path within a window and enables parallel processing of local subsequences, improving efficiency while preserving rich recurrent dynamics. The resulting slice outputs are summarized using the start, end, and mean representations of each slice and then passed to a second stage fixed mixing reservoir for the same direction. This second stage mixes information across slices, allowing the model to combine fine-grained token-level dynamics with broader intra-window context. Its outputs are expanded back to the token level and fused residually with the first stage outputs, which helps preserve detailed local responses while injecting slice-level contextual information.

After slice-level mixing, the four directional outputs are inverse-mapped to a common row-major spatial layout and averaged. This aggregation reduces directional bias and yields a unified window representation for each window. Consecutive blocks alternate between regular and shifted window partitioning. Regular windows support efficient localized processing, whereas shifted windows group tokens that were previously separated, allowing neighboring windows to communicate without resorting to global token mixing and thereby improving cross-window interaction at low cost \cite{liu2021swin}. Finally, a residual feed-forward sublayer with LayerNorm refines these features before global pooling and classification.

Overall, the proposed framework occupies a middle ground between fully trainable modern vision backbones and classical reservoir methods. Unlike attention-based token mixers, it relies on fixed recurrent dynamics for intra-window token interactions, reducing the number of trainable parameters and making the design naturally compatible with reservoir-inspired hardware implementations. Unlike conventional reservoir pipelines that flatten the entire image, it preserves the two-dimensional patch token structure and explicitly incorporates directional spatial traversal together with shifted window communication. In this way, the model combines parameter-efficient recurrent processing with structured modeling of both local regions and their neighbors.

\noindent The main contributions of this work are as follows:
\begin{itemize}
    \item We propose an efficient patch-based image classification architecture that couples multi-directional token scanning with hierarchical reservoir computing, using alternating regular and shifted windows to support both local processing and low-cost cross-window interaction.
    \item We introduce a four-directional window-wise scanning strategy that captures complementary spatial dependencies via horizontal and vertical forward and reverse traversals.
    \item We develop a hierarchical slice-mixing reservoir module in which each directional sequence is first processed by slice-level fixed sub-reservoirs and then integrated by a second-stage fixed mixing reservoir operating on compact slice summaries.
    \item We combine trainable closed-loop readouts, residual slice-level fusion, and a lightweight feed-forward refinement stage to enable end-to-end learning while keeping fixed recurrent dynamics as the primary token mixing mechanism.
\end{itemize}

\section{Related Work}


Image classification has long been dominated by convolutional neural networks (CNNs), whose locality, weight sharing, and translation-equivariant structure provide strong inductive biases for visual recognition. CNNs have advanced hierarchical feature extraction, residual learning, and parameter efficiency, making them important baselines for accuracy--efficiency analysis \cite{lecun2002gradient,he2016deep,iandola2016squeezenet}.

More recent work has expanded visual modeling beyond convolution. Vision Transformer (ViT) showed that images can be represented as sequences of patch tokens and processed effectively using transformer blocks \cite{dosovitskiy2021image}. Swin Transformer further introduced shifted-window processing, enabling local computation together with efficient cross-window interaction \cite{liu2021swin}. Other alternatives to full self-attention have also been explored. MLP-Mixer demonstrated that repeated token-mixing and channel-mixing multilayer perceptrons can be effective for image recognition \cite{tolstikhin2021mlp}, while selective state-space models such as Mamba and VMamba highlighted efficient sequence modeling and multi-directional scanning as promising non-attentional design choices \cite{gu2024mamba, liu2024vmamba}. Recent vision models also employ token-scanning mechanisms to better capture local spatial dependencies by controlling the order and scope of patch interactions, which is especially useful in hierarchical architectures, where accurate local modeling supports progressively richer representations at deeper stages. \cite{pereira2024reviewtransformerbasedmodelscomputer, rahman2024mambavisioncomprehensivesurvey}.

Reservoir computing offers a distinct perspective on sequence modeling. Rather than learning the full recurrent dynamics, reservoir-based methods keep the recurrent core fixed and train only a lightweight readout layer, reducing optimization complexity while preserving nonlinear dynamical processing capability \cite{jaeger2001echo, lukovsevivcius2009reservoir, lukovsevivcius2012practical}. Prior work further showed that deeper or hierarchical reservoir organizations can enhance representational capacity even when recurrent weights remain untrained \cite{gallicchio2017deep}. Extending this idea to vision, ViR demonstrated that images can be represented as patch sequences and processed through a fixed reservoir-based backbone, offering a more parameter- and memory-efficient alternative to fully trainable transformer-style architectures while remaining effective for image classification \cite{wei2021virthevisionreservoir}. Together, these observations motivate the use of structured multi-stage reservoir designs for richer visual representation learning.

Despite these advantages, reservoir computing has been less explored in modern image classification settings. Prior reservoir-based image models often rely on flattened image inputs or simple spatial-to-temporal reformulations, which do not preserve explicit patch-grid structure. For example, multi-reservoir ESN approaches have shown that parallel reservoirs can improve image classification performance, but they generally do not incorporate patch-token embeddings, localized window processing, shifted-window interaction, or directional traversal in the style of contemporary visual backbones \cite{lopez2024exploring}.

Our work addresses these limitations by combining patch-based image tokenization, alternating regular and shifted window partitioning, four-directional local scanning, and hierarchical fixed-reservoir mixing within each window. In contrast to prior reservoir-based image classifiers, the proposed model preserves two-dimensional patch structure, enables efficient cross-window information exchange, and models both fine token-level and coarser slice-level dependencies through directional hierarchical reservoirs. In this way, it connects modern window-based visual modeling with the efficiency of fixed recurrent dynamics \cite{liu2021swin, lopez2024exploring, gu2024mamba, liu2024vmamba}.

\section{HiRo Framework}

\subsection{Overview}

The proposed model is a patch-based image classifier that couples local window processing with directional reservoir computation to perform efficient token mixing. Given an input image, the model first extracts non-overlapping patches, projects them into an embedding space, and augments them with a fixed 2D sinusoidal positional embedding. The resulting sequence of patch tokens is then processed by a stack of Windowed Hierarchical Reservoir (WHR) blocks that jointly model local structure and cross-window interactions.

\begin{figure}[t]
    \centering
    \includegraphics[width=0.90\columnwidth]{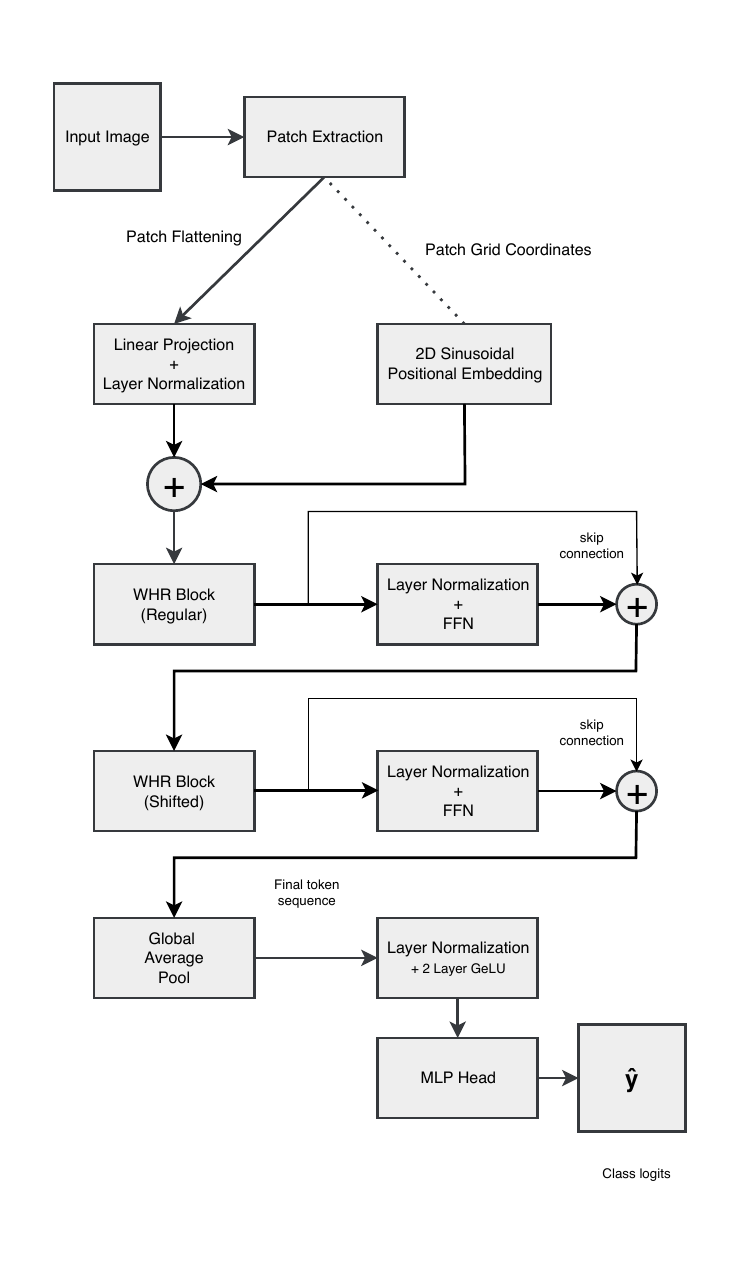}
    \caption{Overview of the proposed directional reservoir-based image classifier.}
    \Description{Pipeline diagram showing patch embedding, positional encoding, regular and shifted window hierarchical reservoir blocks, global pooling, and classifier output.}
    \label{fig:architecture}
\end{figure}

Each WHR block operates on local windows and applies four directional scans within each window: left-to-right, right-to-left, top-to-bottom, and bottom-to-top. These directional sequences are fed into a two-stage hierarchical reservoir module, where slice-level summaries produced by the first stage are further mixed in the second stage and then broadcast back to refine token-level features. The directional outputs are realigned, averaged, merged back into the full token grid, and refined by a residual feed-forward sublayer. The first WHR block uses regular windows, while the second uses shifted windows to enable cross-window interaction at low cost. After the stacked WHR blocks, the final token features are globally averaged and passed to a lightweight classifier head. Figure~\ref{fig:architecture} summarizes the full pipeline. The architectural components are detailed in the following subsections.

\subsection{Patching and Embedding}

We first convert each image into a sequence of patch tokens. Let the input image batch be denoted by \(x\), where \(B\) is the batch size, \(C\) is the number of input channels, and \(H\) and \(W\) are the image height and width, respectively. The model partitions each image into non-overlapping \(P \times P\) patches using tensor unfolding along the spatial dimensions. The extracted patches are then rearranged into a row-major patch sequence of length \(T=(H/P)(W/P)\), so that each patch becomes one token.

Each patch is flattened into a vector of length \(P^{2}C\) and projected by a learned linear layer into a \(d\)-dimensional embedding space. A LayerNorm operation is applied immediately after this projection, producing the embedded token sequence \(\mathbf{u}_{\mathrm{seq}} \in \mathbb{R}^{B \times T \times d}\).

To encode spatial position without adding extra trainable parameters, the embedded tokens are augmented with a fixed 2D sinusoidal positional embedding defined on the patch grid, following the use of sinusoidal positional encodings in sequence models and their adaptation to vision transformers \cite{vaswani2017attention, he2022mae}. Separate sinusoidal encodings are generated for the row and column coordinates, concatenated to form a \(d\)-dimensional positional descriptor for each patch location, and stored as a non-trainable tensor of shape \(1 \times T \times d\). This positional embedding is added elementwise to the patch embeddings before the token sequence is passed to the stacked WHR blocks. In this way, the embedding stage captures both patch content and spatial location.

\subsection{Window Partitioning}

To balance locality and cross-region communication, the architecture employs two WHR blocks with fixed window configurations. Let the token sequence be reshaped into a square token grid \(\mathbf{X} \in \mathbb{R}^{B \times G \times G \times d}\), where \(B\) is the batch size, \(G\) denotes the patch-grid resolution, and \(d\) is the embedding dimension. Let \(W_s\) denote the window size and \(s\) denote the window shift. The first WHR block uses regular non-overlapping \(W_s \times W_s\) windows with \(s=0\), so tokens interact only with other tokens inside the same local window.

The second WHR block introduces cross-communication through shifted windows with
\[
s=\left\lfloor \frac{W_s}{2} \right\rfloor .
\]
Specifically, the token grid is cyclically shifted along both spatial axes by \((-s,-s)\), after which the same non-overlapping \(W_s \times W_s\) partition is applied. Because the shift is cyclic, tokens that were previously separated by regular-window boundaries can be reassigned to the same shifted window under wrap-around indexing. This enables cross-window interaction while preserving localized computation and keeping the window size unchanged. After directional reservoir processing and window reversal, the inverse cyclic shift \((+s,+s)\) restores the original token-grid alignment.

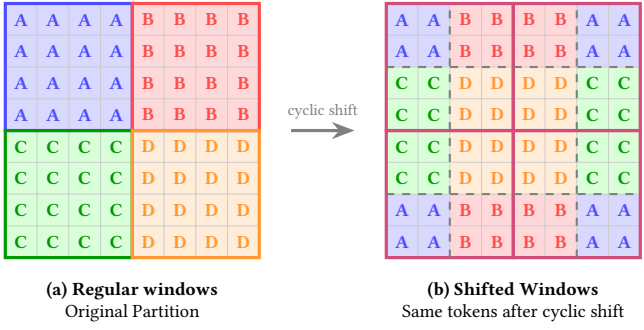
\begin{figure}[t]
  \centering
  \begin{tikzpicture}[scale=0.42, every node/.style={font=\footnotesize}]

    \begin{scope}
      \fill[blue!15]   (0,4) rectangle (4,8);
      \fill[red!15]    (4,4) rectangle (8,8);
      \fill[green!15]  (0,0) rectangle (4,4);
      \fill[orange!15] (4,0) rectangle (8,4);

      \draw[step=1, gray!40, very thin] (0,0) grid (8,8);

      \draw[blue!70,        very thick] (0,4) rectangle (4,8);
      \draw[red!70,         very thick] (4,4) rectangle (8,8);
      \draw[green!60!black, very thick] (0,0) rectangle (4,4);
      \draw[orange!80,      very thick] (4,0) rectangle (8,4);

      \foreach \x in {0,...,3} {
        \foreach \y in {4,...,7} {
          \node[blue!70, font=\footnotesize\bfseries] at (\x+0.5,\y+0.5) {A};
        }
      }
      \foreach \x in {4,...,7} {
        \foreach \y in {4,...,7} {
          \node[red!70, font=\footnotesize\bfseries] at (\x+0.5,\y+0.5) {B};
        }
      }
      \foreach \x in {0,...,3} {
        \foreach \y in {0,...,3} {
          \node[green!60!black, font=\footnotesize\bfseries] at (\x+0.5,\y+0.5) {C};
        }
      }
      \foreach \x in {4,...,7} {
        \foreach \y in {0,...,3} {
          \node[orange!80, font=\footnotesize\bfseries] at (\x+0.5,\y+0.5) {D};
        }
      }

      \node[below] at (4,-0.5) {\textbf{(a) Regular windows}};
      \node[below] at (4,-1.2) {Original Partition};
    \end{scope}

    \draw[-{Stealth[length=3mm]}, very thick, gray] (9,4) -- (11,4);
    \node[above, font=\scriptsize, gray] at (10,4.1) {cyclic shift};

    \begin{scope}[xshift=12cm]

      \foreach \x in {0,1,6,7} {
        \foreach \y in {0,1,6,7} {
          \fill[blue!15] (\x,\y) rectangle (\x+1,\y+1);
        }
      }
      \foreach \x in {2,...,5} {
        \foreach \y in {0,1,6,7} {
          \fill[red!15] (\x,\y) rectangle (\x+1,\y+1);
        }
      }
      \foreach \x in {0,1,6,7} {
        \foreach \y in {2,...,5} {
          \fill[green!15] (\x,\y) rectangle (\x+1,\y+1);
        }
      }
      \foreach \x in {2,...,5} {
        \foreach \y in {2,...,5} {
          \fill[orange!15] (\x,\y) rectangle (\x+1,\y+1);
        }
      }

      \draw[step=1, gray!40, very thin] (0,0) grid (8,8);

      \foreach \x in {0,1,6,7} {
        \foreach \y in {0,1,6,7} {
          \node[blue!70, font=\footnotesize\bfseries] at (\x+0.5,\y+0.5) {A};
        }
      }
      \foreach \x in {2,...,5} {
        \foreach \y in {0,1,6,7} {
          \node[red!70, font=\footnotesize\bfseries] at (\x+0.5,\y+0.5) {B};
        }
      }
      \foreach \x in {0,1,6,7} {
        \foreach \y in {2,...,5} {
          \node[green!60!black, font=\footnotesize\bfseries] at (\x+0.5,\y+0.5) {C};
        }
      }
      \foreach \x in {2,...,5} {
        \foreach \y in {2,...,5} {
          \node[orange!80, font=\footnotesize\bfseries] at (\x+0.5,\y+0.5) {D};
        }
      }

      \draw[purple!70, very thick] (0,4) rectangle (4,8);
      \draw[purple!70, very thick] (4,4) rectangle (8,8);
      \draw[purple!70, very thick] (0,0) rectangle (4,4);
      \draw[purple!70, very thick] (4,0) rectangle (8,4);

      \draw[dashed, gray, thick] (2,0) -- (2,8);
      \draw[dashed, gray, thick] (6,0) -- (6,8);
      \draw[dashed, gray, thick] (0,2) -- (8,2);
      \draw[dashed, gray, thick] (0,6) -- (8,6);

      \node[below] at (4,-0.5) {\textbf{(b) Shifted Windows}};
      \node[below] at (4,-1.2) {Same tokens after cyclic shift};
    \end{scope}
  \end{tikzpicture}
  \caption{Window partitioning on an \(8 \times 8\) token grid (image omitted; only partitions shown). In the regular case, tokens form non-overlapping local windows. In the shifted case, the grid is cyclically shifted before the same partition, grouping tokens near former window boundaries and enabling cross-window interaction while preserving local computation. The shifted layout is expressed in the original grid coordinates, so boundary strips and corners are wrapped portions of full shifted windows.}
  \Description{Two token-grid diagrams comparing regular non-overlapping windows with shifted windows after cyclic shift.}
  \label{fig:window}
\end{figure}

As illustrated in Fig.~\ref{fig:window}, the shifted-window view is shown in the original token-grid coordinates after accounting for cyclic wrap-around. Thus, the side strips and corner regions in the figure are not separate smaller windows, but wrapped portions of full \(W_s \times W_s\) shifted windows when expressed back in the original grid layout. In the present architecture, the first WHR block uses regular windows, and the second uses shifted windows. This shifted-window strategy follows the design introduced in Swin Transformer \cite{liu2021swin}, where alternating regular and shifted window partitioning enables cross-window interaction while maintaining localized computation.

\subsection{Directional Scanning}

Within each local window, we convert the 2D token grid into multiple 1D sequences along four fixed scan directions so that the reservoirs can exploit directional structure. Let a window contain \(W_s \times W_s\) tokens arranged as a local feature grid \(\mathbf{U}_{\mathrm{win}}\). Four ordered sequences are constructed from this grid: left-to-right, right-to-left, top-to-bottom, and bottom-to-top. The left-to-right sequence follows the standard row-major order. The right-to-left sequence is obtained by reversing the column order before flattening. The top-to-bottom sequence is formed by transposing the spatial axes before flattening, and the bottom-to-top sequence applies the same transpose followed by a column reversal.

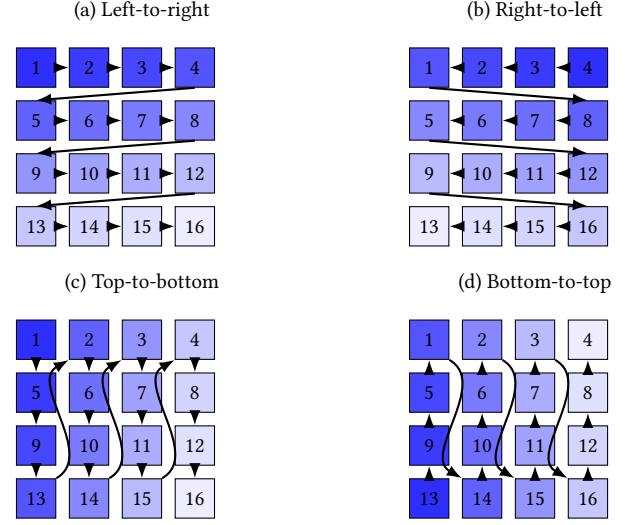
\begin{figure}[t]
\centering
\small
\begin{tikzpicture}[
    cell/.style={draw, minimum width=0.52cm, minimum height=0.52cm, align=center},
    lbl/.style={font=\small},
    arr/.style={-{Latex[length=2mm]}, line width=0.8pt}
]

\colorlet{c1}{blue!82}
\colorlet{c2}{blue!77}
\colorlet{c3}{blue!72}
\colorlet{c4}{blue!67}
\colorlet{c5}{blue!62}
\colorlet{c6}{blue!57}
\colorlet{c7}{blue!52}
\colorlet{c8}{blue!47}
\colorlet{c9}{blue!42}
\colorlet{c10}{blue!37}
\colorlet{c11}{blue!32}
\colorlet{c12}{blue!27}
\colorlet{c13}{blue!22}
\colorlet{c14}{blue!17}
\colorlet{c15}{blue!12}
\colorlet{c16}{blue!7}

\node[lbl] at (1.4,4.55) {(a) Left-to-right};

\node[cell, fill=c1]  (a11) at (0.0,3.8) {1};
\node[cell, fill=c2]  (a12) at (0.7,3.8) {2};
\node[cell, fill=c3]  (a13) at (1.4,3.8) {3};
\node[cell, fill=c4]  (a14) at (2.1,3.8) {4};

\node[cell, fill=c5]  (a21) at (0.0,3.1) {5};
\node[cell, fill=c6]  (a22) at (0.7,3.1) {6};
\node[cell, fill=c7]  (a23) at (1.4,3.1) {7};
\node[cell, fill=c8]  (a24) at (2.1,3.1) {8};

\node[cell, fill=c9]  (a31) at (0.0,2.4) {9};
\node[cell, fill=c10] (a32) at (0.7,2.4) {10};
\node[cell, fill=c11] (a33) at (1.4,2.4) {11};
\node[cell, fill=c12] (a34) at (2.1,2.4) {12};

\node[cell, fill=c13] (a41) at (0.0,1.7) {13};
\node[cell, fill=c14] (a42) at (0.7,1.7) {14};
\node[cell, fill=c15] (a43) at (1.4,1.7) {15};
\node[cell, fill=c16] (a44) at (2.1,1.7) {16};

\draw[arr] (a11.east) -- (a12.west);
\draw[arr] (a12.east) -- (a13.west);
\draw[arr] (a13.east) -- (a14.west);
\draw[arr] (a14.south) -- (a21.north);
\draw[arr] (a21.east) -- (a22.west);
\draw[arr] (a22.east) -- (a23.west);
\draw[arr] (a23.east) -- (a24.west);
\draw[arr] (a24.south) -- (a31.north);
\draw[arr] (a31.east) -- (a32.west);
\draw[arr] (a32.east) -- (a33.west);
\draw[arr] (a33.east) -- (a34.west);
\draw[arr] (a34.south) -- (a41.north);
\draw[arr] (a41.east) -- (a42.west);
\draw[arr] (a42.east) -- (a43.west);
\draw[arr] (a43.east) -- (a44.west);

\node[lbl] at (6.6,4.55) {(b) Right-to-left};

\node[cell, fill=c4]  (b11) at (5.2,3.8) {1};
\node[cell, fill=c3]  (b12) at (5.9,3.8) {2};
\node[cell, fill=c2]  (b13) at (6.6,3.8) {3};
\node[cell, fill=c1]  (b14) at (7.3,3.8) {4};

\node[cell, fill=c8]  (b21) at (5.2,3.1) {5};
\node[cell, fill=c7]  (b22) at (5.9,3.1) {6};
\node[cell, fill=c6]  (b23) at (6.6,3.1) {7};
\node[cell, fill=c5]  (b24) at (7.3,3.1) {8};

\node[cell, fill=c12] (b31) at (5.2,2.4) {9};
\node[cell, fill=c11] (b32) at (5.9,2.4) {10};
\node[cell, fill=c10] (b33) at (6.6,2.4) {11};
\node[cell, fill=c9]  (b34) at (7.3,2.4) {12};

\node[cell, fill=c16] (b41) at (5.2,1.7) {13};
\node[cell, fill=c15] (b42) at (5.9,1.7) {14};
\node[cell, fill=c14] (b43) at (6.6,1.7) {15};
\node[cell, fill=c13] (b44) at (7.3,1.7) {16};

\draw[arr] (b14.west) -- (b13.east);
\draw[arr] (b13.west) -- (b12.east);
\draw[arr] (b12.west) -- (b11.east);
\draw[arr] (b11.south) -- (b24.north);
\draw[arr] (b24.west) -- (b23.east);
\draw[arr] (b23.west) -- (b22.east);
\draw[arr] (b22.west) -- (b21.east);
\draw[arr] (b21.south) -- (b34.north);
\draw[arr] (b34.west) -- (b33.east);
\draw[arr] (b33.west) -- (b32.east);
\draw[arr] (b32.west) -- (b31.east);
\draw[arr] (b31.south) -- (b44.north);
\draw[arr] (b44.west) -- (b43.east);
\draw[arr] (b43.west) -- (b42.east);
\draw[arr] (b42.west) -- (b41.east);

\node[lbl] at (1.4,0.95) {(c) Top-to-bottom};

\node[cell, fill=c1]  (c11) at (0.0,0.2) {1};
\node[cell, fill=c5]  (c12) at (0.7,0.2) {2};
\node[cell, fill=c9]  (c13) at (1.4,0.2) {3};
\node[cell, fill=c13] (c14) at (2.1,0.2) {4};

\node[cell, fill=c2]  (c21) at (0.0,-0.5) {5};
\node[cell, fill=c6]  (c22) at (0.7,-0.5) {6};
\node[cell, fill=c10] (c23) at (1.4,-0.5) {7};
\node[cell, fill=c14] (c24) at (2.1,-0.5) {8};

\node[cell, fill=c3]  (c31) at (0.0,-1.2) {9};
\node[cell, fill=c7]  (c32) at (0.7,-1.2) {10};
\node[cell, fill=c11] (c33) at (1.4,-1.2) {11};
\node[cell, fill=c15] (c34) at (2.1,-1.2) {12};

\node[cell, fill=c4]  (c41) at (0.0,-1.9) {13};
\node[cell, fill=c8]  (c42) at (0.7,-1.9) {14};
\node[cell, fill=c12] (c43) at (1.4,-1.9) {15};
\node[cell, fill=c16] (c44) at (2.1,-1.9) {16};

\draw[arr] (c11.south) -- (c21.north);
\draw[arr] (c21.south) -- (c31.north);
\draw[arr] (c31.south) -- (c41.north);
\draw[arr] (c41.north east) to[out=30,in=210] (c12.south west);
\draw[arr] (c12.south) -- (c22.north);
\draw[arr] (c22.south) -- (c32.north);
\draw[arr] (c32.south) -- (c42.north);
\draw[arr] (c42.north east) to[out=30,in=210] (c13.south west);
\draw[arr] (c13.south) -- (c23.north);
\draw[arr] (c23.south) -- (c33.north);
\draw[arr] (c33.south) -- (c43.north);
\draw[arr] (c43.north east) to[out=30,in=210] (c14.south west);
\draw[arr] (c14.south) -- (c24.north);
\draw[arr] (c24.south) -- (c34.north);
\draw[arr] (c34.south) -- (c44.north);

\node[lbl] at (6.6,0.95) {(d) Bottom-to-top};

\node[cell, fill=c4]  (d11) at (5.2,0.2) {1};
\node[cell, fill=c8]  (d12) at (5.9,0.2) {2};
\node[cell, fill=c12] (d13) at (6.6,0.2) {3};
\node[cell, fill=c16] (d14) at (7.3,0.2) {4};

\node[cell, fill=c3]  (d21) at (5.2,-0.5) {5};
\node[cell, fill=c7]  (d22) at (5.9,-0.5) {6};
\node[cell, fill=c11] (d23) at (6.6,-0.5) {7};
\node[cell, fill=c15] (d24) at (7.3,-0.5) {8};

\node[cell, fill=c2]  (d31) at (5.2,-1.2) {9};
\node[cell, fill=c6]  (d32) at (5.9,-1.2) {10};
\node[cell, fill=c10] (d33) at (6.6,-1.2) {11};
\node[cell, fill=c14] (d34) at (7.3,-1.2) {12};

\node[cell, fill=c1]  (d41) at (5.2,-1.9) {13};
\node[cell, fill=c5]  (d42) at (5.9,-1.9) {14};
\node[cell, fill=c9]  (d43) at (6.6,-1.9) {15};
\node[cell, fill=c13] (d44) at (7.3,-1.9) {16};

\draw[arr] (d41.north) -- (d31.south);
\draw[arr] (d31.north) -- (d21.south);
\draw[arr] (d21.north) -- (d11.south);
\draw[arr] (d11.south east) to[out=-30,in=150] (d42.north west);
\draw[arr] (d42.north) -- (d32.south);
\draw[arr] (d32.north) -- (d22.south);
\draw[arr] (d22.north) -- (d12.south);
\draw[arr] (d12.south east) to[out=-30,in=150] (d43.north west);
\draw[arr] (d43.north) -- (d33.south);
\draw[arr] (d33.north) -- (d23.south);
\draw[arr] (d23.north) -- (d13.south);
\draw[arr] (d13.south east) to[out=-30,in=150] (d44.north west);
\draw[arr] (d44.north) -- (d34.south);
\draw[arr] (d34.north) -- (d24.south);
\draw[arr] (d24.north) -- (d14.south);

\end{tikzpicture}
\caption{Directional scanning within a local \(4 \times 4\) window in the CIFAR setting (image omitted, only token indices shown), illustrating four traversals: left-to-right, right-to-left, top-to-bottom, and bottom-to-top. The same directional ordering is used in both regular and shifted windows; only window membership changes. For MNIST, the window size 
2 yields 4 tokens per local window.}
\Description{Four token-index diagrams showing left-to-right, right-to-left, top-to-bottom, and bottom-to-top scan orders within a local window.}
\label{fig:directional_scanning}
\end{figure}

These four directional sequences provide complementary traversals of the same local content while preserving a deterministic ordering within each window. As illustrated in Fig.~\ref{fig:directional_scanning}, they correspond to horizontal forward, horizontal reverse, vertical forward, and vertical reverse scans. The same directional serialization is used for both regular and shifted windows; only the set of tokens assigned to each window changes.

\subsection{Windowed Hierarchical Reservoir (WHR) Block}

\begin{figure}[t]
    \centering
    \includegraphics[width=\columnwidth]{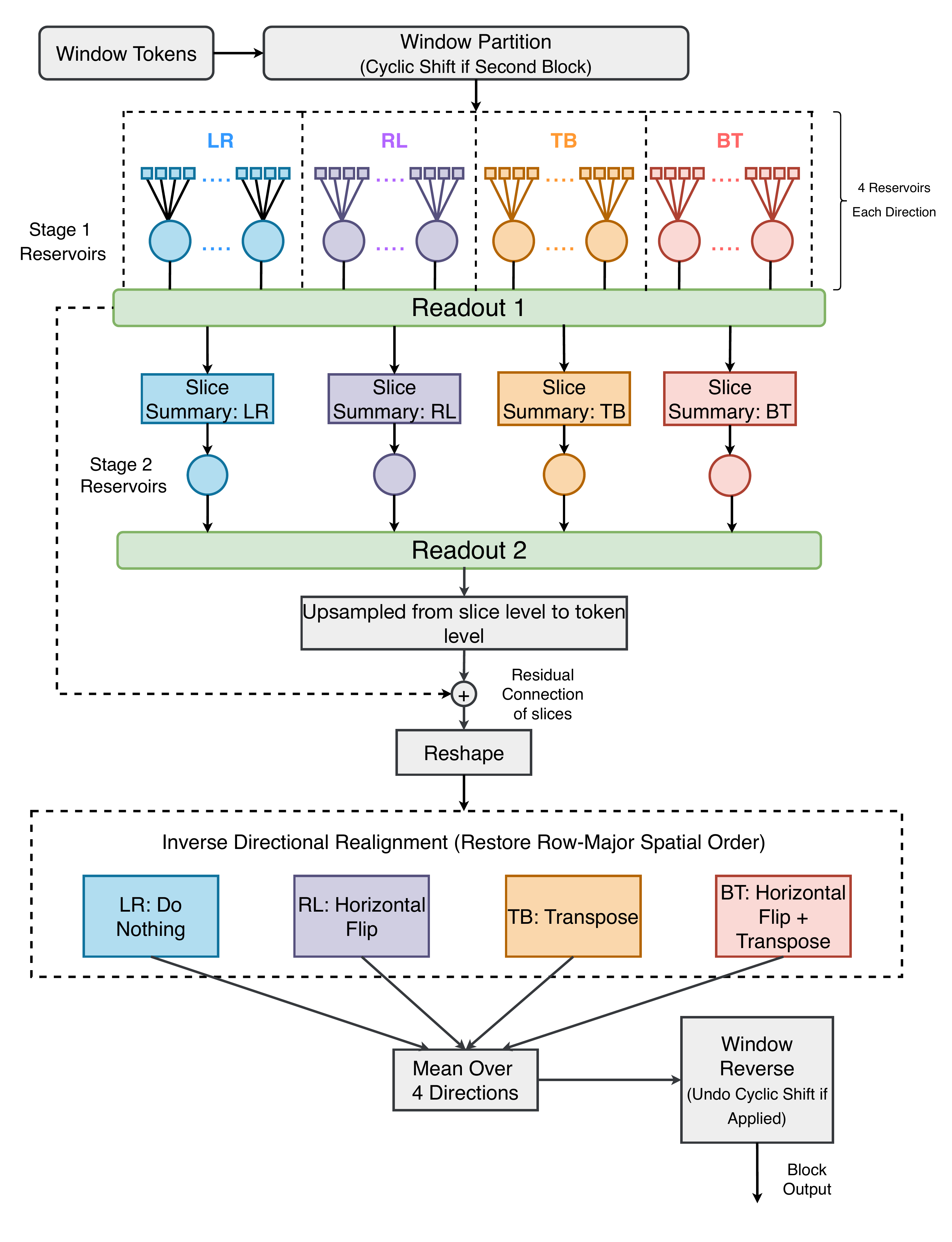}
    \caption{Architecture of a single Windowed Hierarchical Reservoir (WHR) block.}
    \Description{Block diagram showing directional sequences, fixed slice reservoirs, slice summaries, a second-stage mixing reservoir, token expansion, directional fusion, and output refinement.}
    \label{fig:whr_block}
\end{figure}

Each WHR block receives the embedded token sequence and processes each local window using the partitioning and directional scanning schemes described above. For a given window, the four directional token sequences are passed to a two-stage hierarchical reservoir module, after which the outputs are restored to the original spatial order, fused across directions, and written back to the token grid, as illustrated in Fig.~\ref{fig:whr_block}.

The reservoir module is composed of fixed recurrent reservoirs with trainable linear readouts, allowing us to keep the recurrent core parameter-free while retaining nonlinear dynamics. In both hierarchy levels, each reservoir has its own fixed sparse input projection and fixed recurrent matrix, while only the readout parameters are learned. Let \(\mathbf{u}_t\), \(\mathbf{x}_t\), and \(\mathbf{y}_t\) denote the input, reservoir state, and output at step \(t\), respectively. The reservoir state is updated as
\[
\mathbf{x}_t=\tanh\!\left(\mathbf{V}_{\mathrm{in}}\mathbf{u}_t+\mathbf{W}_{\mathrm{res}}\mathbf{x}_{t-1}+\boldsymbol{\epsilon}_t\right),
\]
where \(\mathbf{V}_{\mathrm{in}}\) is the sparse input projection, \(\mathbf{W}_{\mathrm{res}}\) is the fixed recurrent matrix, and \(\boldsymbol{\epsilon}_t\) denotes optional Gaussian noise. The recurrent topology combines bulk, cycle, and long-range jump connections, and the recurrent weights are scaled to control the spectral radius. At the reservoir level, the proposed model uses a structured fixed recurrent topology with bulk, cycle, and long-range jump connections, which is most closely related to cycle-reservoir-with-jumps (CRJ) style designs and to the CRJ-inspired nearly fully connected reservoir used in ViR \cite{rodan2012simple, wei2021virthevisionreservoir}. However, the proposed architecture differs substantially in how this reservoir mechanism is deployed, namely through localized window partitioning, shifted-window interaction, four-directional scanning, and hierarchical slice-level mixing. A short zero-input warm-up is also applied before processing the actual sequence.

The readout design is closely related to that used in ViR \cite{wei2021virthevisionreservoir}, as it applies a trainable linear map to a concatenation of the current input, current reservoir state, previous output, and corresponding squared terms. At each time step \(t\), it forms a feature vector \(\boldsymbol{\phi}_t\) by concatenating three sources of information: the current input \(\mathbf{u}_t\), the current reservoir state \(\mathbf{x}_t\), and the previous output feedback \(\tilde{\mathbf{y}}_{t-1}\). In addition, the readout also includes elementwise squared versions of these terms, scaled by \(g\), in order to introduce simple second-order nonlinear feature interactions. Formally, the readout feature is defined as
\[
\boldsymbol{\phi}_t=
\big[
\mathbf{u}_t,\,
\mathbf{x}_t,\,
\tilde{\mathbf{y}}_{t-1},\,
g\,\mathbf{u}_t^{\odot 2},\,
g\,\mathbf{x}_t^{\odot 2},\,
g\,\tilde{\mathbf{y}}_{t-1}^{\odot 2}
\big],
\]
and the current output is obtained by a trainable linear map,
\[
\mathbf{y}_t=\mathbf{R}\boldsymbol{\phi}_t+\mathbf{b},
\]
where \(g\) is a scaling factor and \(\tilde{\mathbf{y}}_{t-1}=\tanh(\mathbf{y}_{t-1})\) when output squashing is used. In this way, the readout can combine immediate input information, the current recurrent memory, and output feedback within a single learned projection.

In the first hierarchy level, each directional sequence of length \(W_s^2\) is divided into \(M\) equal contiguous slices. All slices from all four directions are then processed in parallel by distinct stage-1 reservoirs that share a common readout. This first stage preserves token-level resolution within each slice. To build a coarser representation for the second stage, each slice is summarized by concatenating three statistics from its stage-1 output: the first token response, the last token response, and the mean response over the slice. For slice \(m\), the summary is
\[
\mathbf{s}_m=
\big[
\mathbf{y}^{(1)}_{m,1}\,\|\,\mathbf{y}^{(1)}_{m,L_s}\,\|\,
\frac{1}{L_s}\sum_{\ell=1}^{L_s}\mathbf{y}^{(1)}_{m,\ell}
\big],
\]
where \(L_s\) is the slice length and \(\|\) denotes concatenation. These ordered slice summaries are then passed, separately for each direction, through a second bank of fixed mixing reservoirs with a different shared readout. Thus, the first stage models fine token-level dynamics, while the second stage mixes information across slices within the same direction.

The stage-2 outputs are produced at slice resolution and are therefore broadcast back to token resolution by repeating each slice representation across the tokens belonging to that slice. This expanded representation is added to the stage-1 token features as a slice residual, allowing coarser directional context to refine the finer token-level responses. The resulting directional outputs are then inverse-aligned to the original row-major geometry and averaged across the four scan directions to form a single feature map for each window.

Finally, the window features are merged back into the full token grid, and the inverse cyclic shift is applied when the block uses shifted windows. The resulting token sequence is then refined by a residual feed-forward sublayer consisting of LayerNorm, a two-layer MLP with hidden dimension \(4d\), and a GELU activation. \textit{In this way, each WHR block combines fixed reservoir dynamics, hierarchical slice-level mixing, slice-to-token context injection, directional fusion, and residual token refinement within a single module}. 

\subsection{Prediction Head}

After the stacked WHR blocks, the final token features are globally averaged across the token dimension to produce a single image-level feature vector of dimension \(d\). This pooled feature is passed through a final LayerNorm and then into a two-layer MLP classifier. The first linear layer maps from \(d\) to \(d\), followed by a GELU activation, and the second linear layer maps from \(d\) to \(K\), where \(K\) is the number of classes. In the experiments, \(K=10\) for MNIST and CIFAR-10, and \(K=100\) for CIFAR-100. The prediction head remains intentionally lightweight because the WHR backbone already performs the main spatial and contextual modeling; it only maps the globally pooled image feature to the final class logits.

\section{Experiments}

\subsection{Experimental Setup and Training Details}
We evaluate the proposed model against convolutional, transformer-based, and reservoir-based baselines under a unified training protocol to enable fair comparison. 
All gradient-trained models were implemented in PyTorch, using official implementations when available and modifying only input and classifier layers for dataset compatibility. Core optimization and training hyperparameters are listed in Table~\ref{tab:hyperparams}. All backpropagation-based models are trained from scratch for 80 epochs on MNIST and 150 epochs on CIFAR-10 and CIFAR-100, with 2 warmup epochs in each setting, and are evaluated using accuracy, macro-averaged precision, recall, and F1-score. To improve robustness and regularization, we add Gaussian noise with standard deviation \(0.02\) to the inputs during training for all gradient-based models, but disable it at test time.

The baseline set spans classical convolutional, lightweight convolutional, transformer, and reservoir-based architectures. LeNet is minimally adapted to the dataset resolution and channel dimensionality: for MNIST, the final \(5\times5\) convolution is reduced to \(4\times4\) to match the native \(28\times28\) resolution, while for CIFAR-10 and CIFAR-100 the original \(32\times32\) configuration is retained. ResNet-18 is adjusted for small-resolution inputs by replacing the default stem with a \(3\times3\), stride-1 convolution and removing the initial max-pooling layer, which avoids excessive early downsampling. ViT (Small/4) is implemented as a compact Vision Transformer with \(4\times4\) patches, 6 transformer layers, 6 attention heads, hidden dimension 384, and MLP dimension 1536; for MNIST, grayscale images are replicated across three channels to match the standard ViT input format. Swin (Small) preserves the standard Swin-Small stage configuration with embedding dimension 96, depths \([2,2,18,2]\), heads \([3,6,12,24]\), and MLP ratio 4.0, but uses smaller patch/window sizes of \(4/7\) for MNIST and \(2/4\) for CIFAR-10 and CIFAR-100 to accommodate low-resolution images. SqueezeNet and the remaining neural baselines are modified only at the input and classification heads when necessary, keeping their internal architectures unchanged.

MRESN \cite{lopez2024exploring} is treated separately because it is not trained end-to-end by backpropagation and therefore cannot be aligned exactly with the gradient-based training setup used for the gradient-trained models. Instead of patchified image tokens and iterative gradient updates, it uses flattened image vectors and a closed-form ridge-regression readout over fixed random reservoirs. We implement MRESN as a multi-reservoir echo state network with 20 parallel reservoirs of 1000 nodes each, reservoir sparsity 0.2, leak rate 0.7, Tikhonov regularization \(10^{-8}\), and a 2-step initial transient. For diversity across reservoirs, spectral radius and input scaling for each reservoir are sampled independently from \([0.5,1.5]\). Inputs are flattened rather than patchified, yielding 784-dimensional inputs for MNIST and 3072-dimensional inputs for CIFAR-10 and CIFAR-100, matching the original formulation.

\begin{table}[t]
\centering
\caption{Experimental settings and key hyperparameters.}
\label{tab:hyperparams}
\setlength{\tabcolsep}{4pt}
\renewcommand{\arraystretch}{1.05}
\begin{tabular}{@{}p{0.52\linewidth}p{0.42\linewidth}@{}}
\toprule
\multicolumn{2}{@{}l@{}}{\textsc{\textbf{Common Training Setup}}} \\
\midrule
Hardware & NVIDIA T4 (Google Colab) \\
Framework & PyTorch \\
Initialization & From scratch (no pretraining) \\
Random seed & 42 \\
Loss function & Cross-entropy \\
Optimizer & AdamW ($\beta_1{=}0.9,\;\beta_2{=}0.999$) \\
Learning rate & $3 \times 10^{-4}$ \\
Weight decay & $1 \times 10^{-5}$ \\
LR schedule & Cosine decay w/ linear warmup \\
Minimum LR & $1 \times 10^{-5}$ \\
Batch size & 128 \\
Gradient clipping & max-norm $= 1.0$ \\
Training noise $\sigma$ & 0.02 (gradient-based only) \\
\midrule
\multicolumn{2}{@{}l@{}}{\textsc{\textbf{MNIST}}} \\
\midrule
Train / val / test split & 55k / 5k / 10k \\
Input resolution & $28 \times 28$ \\
Epochs / warmup epochs & 80 / 2 \\
Augmentation & Random crop (pad=2) \\
Patch / window / slices & $7$ / $2$ / $4$ \\
Reservoir dimension $d_r$ & 200 \\
\midrule
\multicolumn{2}{@{}l@{}}{\textsc{\textbf{CIFAR-10 \& CIFAR-100}}} \\
\midrule
Train / val / test split & 45k / 5k / 10k \\
Input resolution & $32 \times 32$ \\
Epochs / warmup epochs & 150 / 2 \\
Augmentation & Random crop (pad=4) + horiz.\ flip \\
Patch / window / slices & $4$ / $4$ / $4$ \\
Reservoir dimension $d_r$ & 200 \\
\midrule
\multicolumn{2}{@{}l@{}}{\textsc{\textbf{Reservoir Configuration}}} \\
\midrule
Bulk / cycle / jump weights & 0.08 / 0.05 / 0.5 \\
Spectral scaling & 0.9 \\
Jump size & 137 \\
$\phi^2$ scale & 0.1 \\
\bottomrule
\end{tabular}
\end{table}

\subsection{Results and Comparison}

\begin{table*}[t]
\centering
\caption{Performance comparison across models on MNIST, CIFAR-10, and CIFAR-100. All metrics are reported in percentage (\%). HiRo results are shown in bold, and the best non-HiRo baseline for each metric is underlined.}
\label{tab:main_results}
\resizebox{0.8\textwidth}{!}{%
\begin{tabular}{lcccccccccccc}
\toprule
\multirow{2}{*}{\textbf{Model}} 
& \multicolumn{4}{c}{\textbf{MNIST}} 
& \multicolumn{4}{c}{\textbf{CIFAR-10}} 
& \multicolumn{4}{c}{\textbf{CIFAR-100}} \\
\cmidrule(lr){2-5} \cmidrule(lr){6-9} \cmidrule(lr){10-13}
& \textbf{Acc.} & \textbf{Prec.} & \textbf{Rec.} & \textbf{F1}
& \textbf{Acc.} & \textbf{Prec.} & \textbf{Rec.} & \textbf{F1}
& \textbf{Acc.} & \textbf{Prec.} & \textbf{Rec.} & \textbf{F1} \\
\midrule
LeNet 
& 99.30 & 99.30 & 99.28 & 99.29
& 62.71 & 62.33 & 62.71 & 62.32
& 31.88 & 30.47 & 31.88 & 30.53 \\
SqueezeNet 
& 99.41 & 99.41 & 99.41 & 99.41
& 68.76 & 62.78 & 68.76 & 65.39
& 18.84 & 7.57  & 18.84 & 10.44 \\
ResNet-18 
& \underline{99.73} & \underline{99.73} & \underline{99.73} & \underline{99.73}
& \underline{93.02} & \underline{93.00} & \underline{93.02} & \underline{93.00}
& \underline{70.54} & \underline{70.60} & \underline{70.54} & \underline{70.42} \\
ViT (Small/4)
& 99.48 & 99.48 & 99.48 & 99.48
& 82.23 & 82.30 & 82.23 & 82.24
& 55.18 & 54.98 & 55.18 & 54.85 \\
Swin (Small) 
& 99.38 & 99.37 & 99.37 & 99.37
& 84.13 & 84.04 & 84.13 & 84.07
& 53.52 & 53.18 & 53.52 & 53.16 \\
MRESN
& 97.70 & 97.68 & 97.69 & 97.68
& 36.32 & 36.29 & 36.32 & 36.22
& 17.41 & 16.51 & 17.41 & 16.71 \\
\midrule
HiRo (Ours) 
& \textbf{99.46} & \textbf{99.47} & \textbf{99.45} & \textbf{99.46}
& \textbf{85.57} & \textbf{85.60} & \textbf{85.57} & \textbf{85.58}
& \textbf{59.10} & \textbf{59.34} & \textbf{59.10} & \textbf{59.03} \\
\bottomrule
\end{tabular}%
}
\end{table*}

\begin{table*}[t]
\centering
\caption{Efficiency and GPU resource comparison across CNN, Transformer, and reservoir computing-based models on MNIST, CIFAR-10, and CIFAR-100. Params denotes trainable parameters in millions (M). Epoch and inference times are reported in seconds. Peak CUDA memory is reported in MB for training and inference. GPU utilization is reported as mean utilization (\%) for training and inference. MRESN is trained in a single closed-form pass without epoch-based backpropagation. HiRo results are shown in bold, and the best non-HiRo baseline for each metric is underlined.}
\label{tab:merged_efficiency_gpu}
\setlength{\tabcolsep}{3pt}
\renewcommand{\arraystretch}{0.95}
\resizebox{0.75\textwidth}{!}{%
\begin{tabular}{llccccccc}
\toprule
\textbf{Dataset} & \textbf{Model} & \textbf{Params} & \textbf{Epoch} & \textbf{Inf.} & \textbf{Peak (Tr.)} & \textbf{Peak (Inf.)} & \textbf{GPU Util. (Tr.)} & \textbf{GPU Util. (Inf.)} \\
\midrule
\multirow{7}{*}{MNIST}
& LeNet         & \underline{0.044} & \underline{21.93} & \underline{2.35} & \underline{30.4}   & \underline{24.0}   & \underline{11.9} & \underline{4.5} \\
& SqueezeNet    & 0.728 & 30.50 & 3.19 & 62.9   & 38.9   & 23.5 & 5.6 \\
& ResNet-18     & 11.173 & 41.55 & 2.83 & 663.8  & 399.3  & 93.5 & 64.8 \\
& ViT (Small/4) & 10.690 & 32.09 & 3.27 & 753.1  & 295.5  & 71.0 & 31.0 \\
& Swin (Small)  & 48.845 & 138.59 & 7.86 & 2356.5 & 943.6  & 92.7 & 79.0 \\
& MRESN         & 0.208 & --    & 2.80 & 229.3  & 229.3  & -  & 29.5 \\
& HiRo (Ours)   & \textbf{0.887} & \textbf{31.84} & \textbf{3.06} & \textbf{692.7}  & \textbf{171.1}  & \textbf{70.1} & \textbf{52.1} \\
\midrule
\multirow{7}{*}{CIFAR-10}
& LeNet         & \underline{0.061} & \underline{21.72} & 3.14 & \underline{47.7}   & \underline{39.4}   & \underline{12.3} & \underline{5.1} \\
& SqueezeNet    & 0.728 & 23.10 & 2.63 & 81.0   & 54.3   & 25.7 & 7.8 \\
& ResNet-18     & 11.174 & 39.42 & 3.18 & 828.0  & 463.4  & 92.1 & 53.5 \\
& ViT (Small/4) & 10.696 & 31.28 & 2.88 & 930.8  & 320.3  & 80.9 & 32.5 \\
& Swin (Small)  & 48.808 & 105.60 & 7.14 & 2269.8 & 933.1  & 94.5 & 82.1 \\
& MRESN         & 0.231 & --    & \underline{1.07} & 373.2  & 373.2  & -  & 50.0 \\
& HiRo (Ours)   & \textbf{0.887} & \textbf{39.40} & \textbf{3.53} & \textbf{1120.7} & \textbf{224.2}  & \textbf{87.7} & \textbf{65.8} \\
\midrule
\multirow{7}{*}{CIFAR-100}
& LeNet         & \underline{0.069} & \underline{21.42} & \underline{2.50} & \underline{49.9}   & \underline{28.2}   & \underline{14.4} & \underline{4.9} \\
& SqueezeNet    & 0.774 & 23.70 & 2.61 & 69.1   & 42.3   & 28.2 & 5.5 \\
& ResNet-18     & 11.220 & 39.94 & 3.01 & 829.5  & 464.7  & 92.6 & 53.4 \\
& ViT (Small/4) & 10.730 & 30.46 & 3.64 & 931.3  & 322.0  & 85.0 & 44.1 \\
& Swin (Small)  & 48.877 & 104.92 & 7.78 & 2273.2 & 932.6  & 95.1 & 84.8 \\
& MRESN         & 2.307 & --    & 1.24 & 373.2  & 373.2  & -  & 41.0 \\
& HiRo (Ours)   & \textbf{0.898} & \textbf{40.84} & \textbf{3.54} & \textbf{1118.8} & \textbf{223.7}  & \textbf{89.1} & \textbf{53.0} \\
\bottomrule
\end{tabular}%
}
\end{table*}


Table~\ref{tab:main_results} summarizes the predictive performance of all models on MNIST, CIFAR-10, and CIFAR-100. On MNIST, all deep models reach very high accuracy and F1, indicating a near-saturated regime in which architectural differences matter little. In this setting, HiRo remains competitive, achieving 99.46\% accuracy and 99.46\% F1, comparable to ViT and Swin and only slightly below ResNet-18. The benefits of HiRo become clearer on the more challenging CIFAR benchmarks. On CIFAR-10, HiRo attains 85.57\% accuracy and 85.58\% F1, outperforming ViT (82.23\%) and Swin (84.13\%) and greatly improving over the reservoir baseline MRESN (36.32\%), while ResNet-18 remains the strongest overall at 93.02\%. On CIFAR-100, HiRo reaches 59.10\% accuracy and 59.03\% F1, again exceeding both transformer baselines (55.18\% for ViT and 53.52\% for Swin) and providing a large gain over MRESN (17.41\%). These results indicate that HiRo preserves the robustness of standard deep architectures on easy data and scales substantially better than a flat multi-reservoir ESN when moving to higher-resolution, multi-class natural-image tasks.


Table~\ref{tab:merged_efficiency_gpu} reports parameter counts, runtime, and GPU resource usage, highlighting HiRo’s accuracy–efficiency trade-off. Across all datasets, HiRo uses fewer than 1M trainable parameters, making it an order of magnitude more compact than ResNet-18 ($\approx$11M) and ViT ($\approx$10.7M), and dramatically smaller than Swin ($\approx$49M), while still matching or outperforming the transformer baselines on CIFAR-10 and CIFAR-100. LeNet and SqueezeNet are smaller, but their performance degrades sharply on the CIFAR datasets, showing that further parameter reductions come at a significant accuracy cost.

Runtime and GPU measurements place HiRo in a practical middle ground between lightweight CNNs and heavy transformer backbones. Its per-epoch training time is comparable to ViT and substantially lower than Swin, despite using richer intra-window reservoir dynamics. In terms of memory, HiRo’s training-time peak usage is higher than that of the smallest CNNs and MRESN, but remains far below Swin and close to ViT; at inference time, HiRo consistently uses less peak memory than ResNet-18 and ViT on all three datasets. GPU utilization during training is also high, especially on CIFAR-10 and CIFAR-100, indicating that the model structure makes effective use of the available hardware. Overall, these results support HiRo as a compact, reservoir-based backbone that delivers transformer-level or better accuracy on natural images at a fraction of the parameter budget and with substantially lower resource demands than the heaviest transformer variant.


\section{Discussion and Future Directions}

Our results show that combining shifted-window processing with directional hierarchical reservoirs yields an effective balance among locality, cross-window interaction, and parameter efficiency. In particular, the model benefits from the complementary roles of these two components: shifted-window partitioning provides structured local processing together with low-cost communication across neighboring windows, while the directional reservoir hierarchy supplies lightweight recurrent token mixing within each window. This combination is especially advantageous on CIFAR-10 and CIFAR-100, where very lightweight baselines lose substantial accuracy and heavier models require much larger parameter budgets \cite{liu2021swin}.

These findings highlight that shifted-window partitioning and directional reservoir computation function as synergistic rather than competing design choices. The window mechanism preserves spatial structure and supports efficient neighborhood interaction, whereas the reservoir hierarchy injects nonlinear recurrent processing over multiple directional traversals. Jointly, they offer a compact alternative to standard trainable token mixers, suggesting that modern vision design principles can be fruitfully combined with fixed dynamical systems \cite{liu2021swin,lukovsevivcius2009reservoir,wei2021virthevisionreservoir}.


Looking ahead, a first natural direction is to test the scalability of HiRo on more complex vision tasks. Extending the framework to higher-resolution image classification, dense prediction, object detection, and semantic segmentation would provide a stronger assessment of whether hierarchical directional reservoirs can move beyond small-image benchmarks while retaining their efficiency advantages.

A second, complementary direction is to exploit the connection between HiRo and physical reservoir computing. The framework is motivated not only by algorithmic efficiency but also by the broader potential of reservoir computing as a bridge to physical dynamical substrates. A particularly promising avenue is to explore physical reservoir computing implementations based on patterned nanomagnet arrays, artificial spin ice, and electro-photonics. Recent work has shown that geometrically frustrated nanomagnetic systems and integrated electro-photonic circuits can realize practical neuromorphic reservoirs with rich nonlinear dynamics, echo-state behavior, and short-term memory \cite{ti2024mixed,tanaka2019recent, allwood2023perspective, vatsavai2021silicon}. Building on this perspective, an important future direction for HiRo is to investigate whether its hierarchical directional reservoir mechanism can be adapted to artificial-spin-ice-based or nanomagnet-array-based hardware, thereby linking compact patch-based vision modeling with physical reservoirs.

\section{Conclusion}

We presented a directional reservoir-based image classification framework that combines patch embeddings, 2D sinusoidal positional encoding, and local window partitioning. Within each window, tokens are processed along four scan directions using a two-stage fixed reservoir hierarchy, composed of slice-level directional reservoirs followed by direction-wise mixing reservoirs, each equipped with a shared linear readout. By replacing fully trainable intra-window token mixing with fixed recurrent dynamics, the proposed model preserves localized spatial processing while enabling cross-window interaction through alternating regular and shifted windows. Our empirical results indicate that reservoir-based token mixing is a promising design choice for compact visual recognition backbones, offering competitive accuracy with substantially fewer trainable parameters than standard transformer-style architectures.

\end{document}